\documentclass[nohyperref]{article}

\usepackage{microtype}
\usepackage{graphicx}
\usepackage{subfigure}
\usepackage{booktabs}
\usepackage{hyperref}
\usepackage{amsmath}
\usepackage{amssymb}
\usepackage{mathtools}
\usepackage{amsthm}
\usepackage[capitalize,noabbrev]{cleveref}
\usepackage{xcolor}
\usepackage{booktabs}
\usepackage{xspace}

\usepackage[accepted]{icml2022}

\newcommand{\method}{MVP\xspace}
\newcommand{\pmc}{PixMC\xspace}
\newcommand{\x}{$\times$}
\newcommand{\app}{\raise.17ex\hbox{$\scriptstyle\sim$}}
\renewcommand{\paragraph}[1]{\textbf{#1}}

\begin{document}
\twocolumn[
\icmltitle{Masked Visual Pre-training for Motor Control}
\icmlsetsymbol{star}{*}
\icmlsetsymbol{starstar}{**}
\icmlsetsymbol{dagger}{$\dagger$}
\begin{icmlauthorlist}
\icmlauthor{Tete Xiao}{star,berkeley}
\icmlauthor{Ilija Radosavovic}{star,berkeley}
\icmlauthor{Trevor Darrell}{dagger,berkeley}
\icmlauthor{Jitendra Malik}{dagger,berkeley}
\end{icmlauthorlist}
\icmlaffiliation{berkeley}{University of California, Berkeley}
\icmlcorrespondingauthor{Tete Xiao}{txiao@eecs.berkeley.edu}
\icmlcorrespondingauthor{Ilija Radosavovic}{ilija@berkeley.edu}
\icmlkeywords{Machine Learning, ICML}
\vskip 0.3in
]
\printAffiliationsAndNotice{\icmlEqualContribution}

\begin{abstract}
This paper shows that self-supervised visual pre-training from real-world images is effective for learning motor control tasks from pixels.
We first train the visual representations by masked modeling of natural images.
We then freeze the visual encoder and train neural network controllers on top with reinforcement learning.
We do not perform any task-specific fine-tuning of the encoder; the same visual representations are used for all motor control tasks.
To the best of our knowledge, this is the first self-supervised model to exploit real-world images at scale for motor control.
To accelerate progress in learning from pixels, we contribute a benchmark suite of hand-designed tasks varying in movements, scenes, and robots.
Without relying on labels, state-estimation, or expert demonstrations, we consistently outperform supervised encoders by up to 80\% absolute success rate, sometimes even matching the oracle state performance.
We also find that in-the-wild images, e.g., from YouTube or Egocentric videos, lead to better visual representations for various manipulation tasks than ImageNet images.
\end{abstract}

\section{Introduction}

The last decade of machine learning has been powered by learning representations with large neural networks and augmenting them with a relatively small amount of domain knowledge where appropriate. This paradigm has led to substantial progress across a range of domains. Examples include visual recognition~\cite{Girshick2014, He2017}, natural language~\cite{Radford2018, Devlin2019, Radford2019, Brown2020}, and audio~\cite{Van2016}. And the trend continues. Motor control, however, remains a notable exception.

\newpage

In this paper, we show that self-supervised visual pre-training on real-world images is effective for learning motor control tasks from pixels. These self-supervised representations consistently outperform supervised representations.

\begin{figure}[t]\centering\vspace{-0mm}
\includegraphics[width=0.98\linewidth]{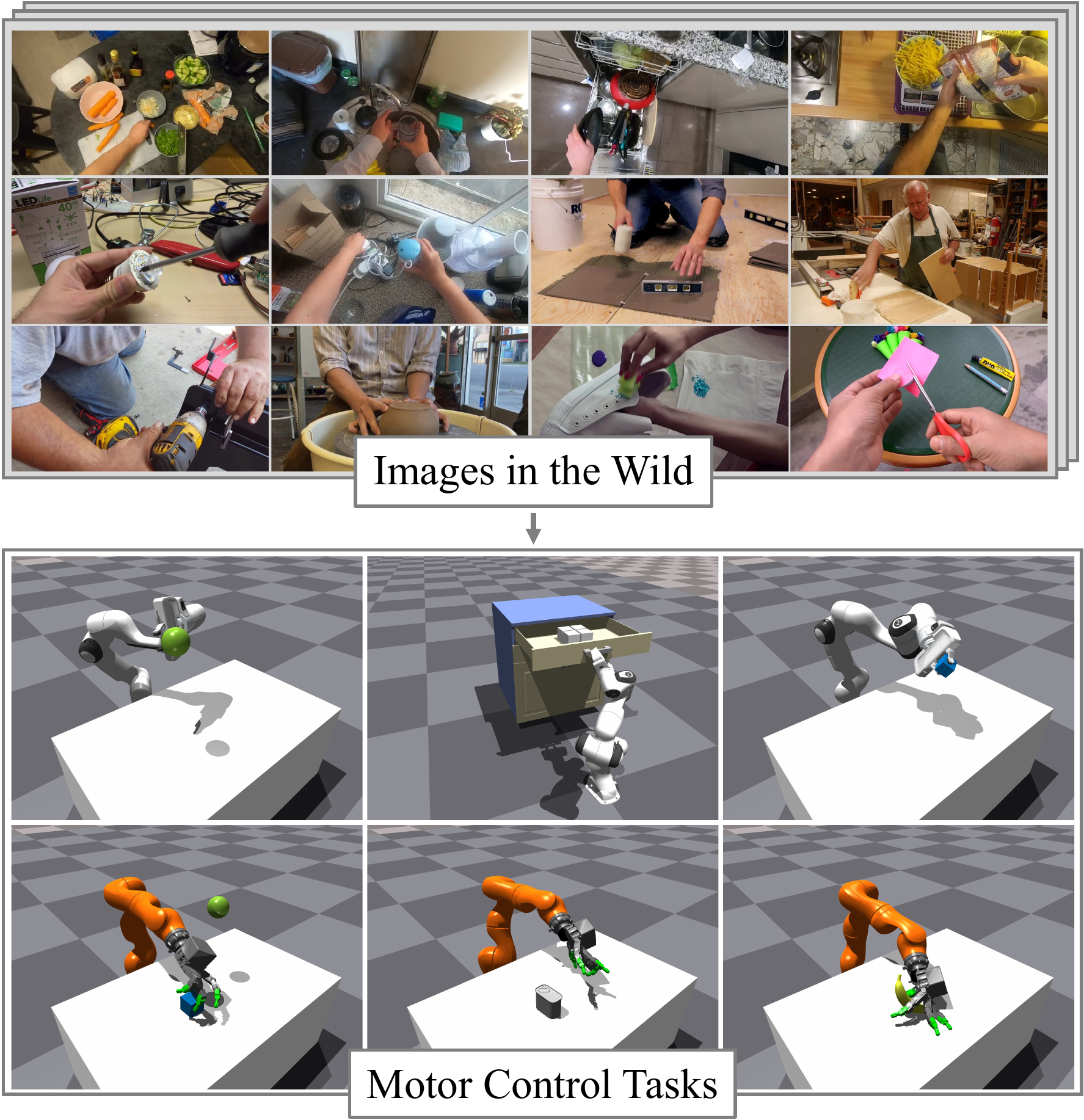}\vspace{-2mm}
\caption{We explore learning complex motor control from pixels. We show that we are able to solve a range of motor control tasks with variations in robots, scenes, and objects. This is enabled by learning rich visual representations from real-world images with masked modeling. Videos available on the \textbf{\href{https://tetexiao.com/projects/mvp}{project page}}.}
\label{fig:teaser}\vspace{-4mm}
\end{figure}

Consider tasks shown in Figure~\ref{fig:teaser} (bottom). The required movement types vary from simple reaching to object interactions. We also see variations in robots, scene configurations, and objects. Control inputs are high-dimensional and difficult to search (e.g., 23 DoF robot with a multi-finger hand). We explore learning complex tasks such as these from high-dimensional pixel observations.

\begin{figure*}[t]\centering\vspace{-0mm}
\includegraphics[width=1.0\linewidth]{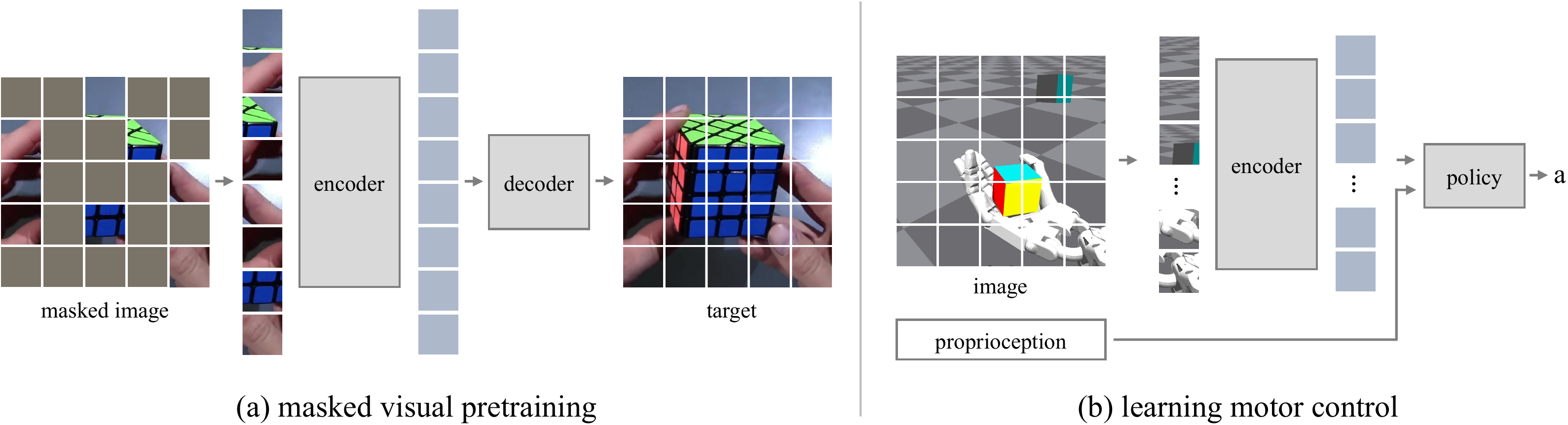}\vspace{-4mm}
\caption{\textbf{Masked visual pre-training for motor control.} \emph{Left:} We first \emph{pre-train} visual representations using \emph{self-supervision} through masked image modeling~\cite{He2021} from \emph{real-world} images. \emph{Right:} We then \emph{freeze} the image encoder and train task-specific controllers on top with reinforcement learning (RL). The \emph{same} visual representations are used for all motor control tasks.}
\label{fig:framework}\vspace{-0mm}
\end{figure*}

To tackle this setting, we use the neural network architecture shown in Figure~\ref{fig:framework}b. Our network encodes the input image using a high-capacity visual encoder~\cite{Dosovitskiy2020} and combines it with proprioceptive information to obtain an embedding. A light-weight neural network controller takes in the embedding and predicts actions. The whole system can be trained end-to-end.

Indeed, this design is akin to architectures typically used in approaches that learn control policies end-to-end with RL, e.g.,~\citet{Levine2016}. While conceptually appealing, the latter has two main challenges in practice. First, training is computationally expensive and has poor sample complexity (especially with high-dimensional inputs and actions). Second, the learned solutions typically overfit to the setting at hand and thus do not generalize to new scenes and objects.

One way to offset the high sample complexity of end-to-end RL is to employ auxiliary objectives~\cite{Jaderberg2016,Oord2018,Yarats2019,Srinivas2020}. For example, \citet{Srinivas2020} show excellent performance in vision-based RL by using contrastive learning with data augmentations. However, such representations are still trained using only environment-specific experience.

The key aspect of our approach is in how we train the visual representations. We do \emph{not} train the visual encoder while learning specific motor control tasks. 
Instead, we \emph{pre-train} the visual encoder by \emph{self-supervision} from \emph{natural images} (Figure~\ref{fig:teaser} \& \ref{fig:framework}).
We learn the visual representation by performing masked image modeling through the masked autoencoder (MAE)~\cite{He2021}. Thanks to the Internet and ubiquitous portable cameras, we now have access to large amounts of unlabeled visual data for self-supervision. MAE does not require human labels or make strong assumptions about data distributions, e.g., centered objects or pre-defined augmentation invariances~\cite{Xiao2021a}, making it an excellent framework for learning general visual representations from large collections of in-the-wild images.

Given the visual encoder, we train controllers on top with reinforcement learning~\cite{Schulman2017}. We keep the visual representations frozen and do not perform \emph{any} task-specific fine-tuning of the encoder; all motor control tasks use the \emph{same} visual representations. We call our approach \method (for \textbf{M}asked \textbf{V}isual \textbf{P}re-training for Motor Control).

To accelerate future progress, we contribute \pmc, a new benchmark suite of hand-designed tasks with various movement types, scene configurations, and robots. We leverage a GPU-based simulator for fast simulation~\cite{Makoviychuk2021}, provide reward functions, baselines, and multi-GPU implementation of learning algorithms from pixels.

\begin{figure*}[t]\centering
\includegraphics[width=0.95\linewidth]{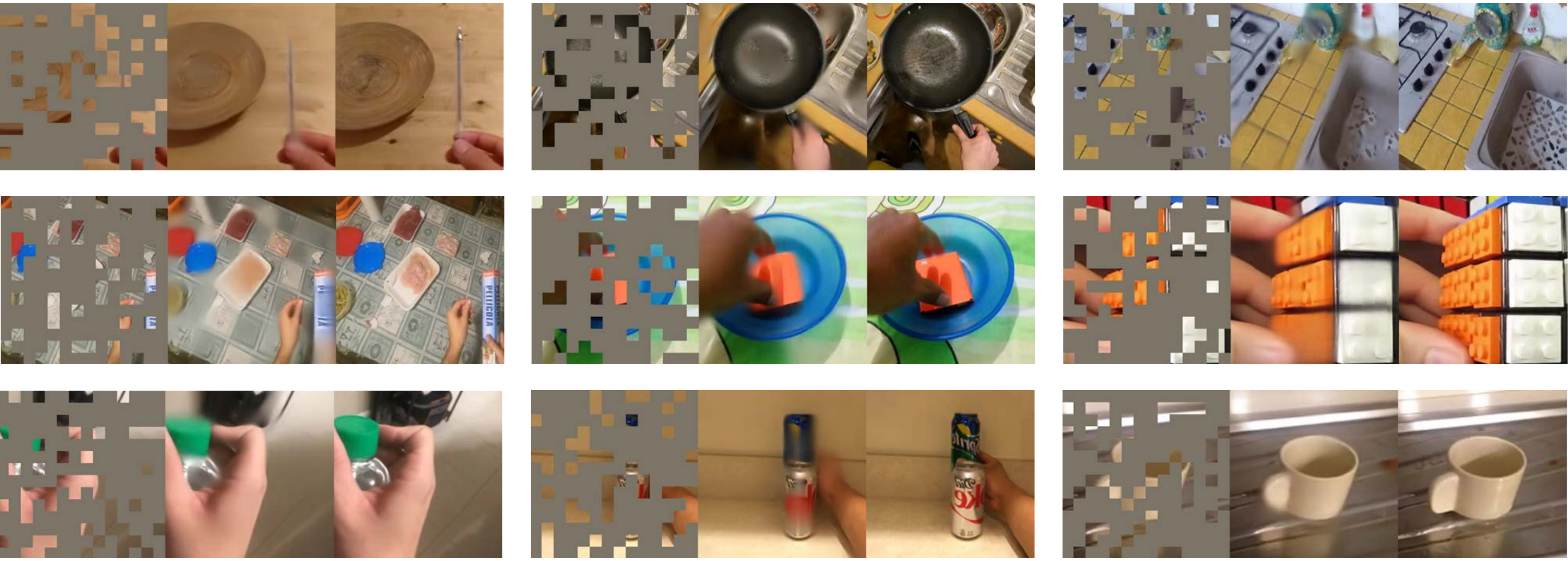}\vspace{-2mm}
\caption{\textbf{Example reconstructions.} For each triplet from left to right: the masked image, the reconstructed image, the ground-truth target. We observe that the autoencoder learns color, shape, and object affordance using self-supervision from in-the-wild images.}
\label{fig:mae_vis}\vspace{-0mm}
\end{figure*}

We compare our self-supervised approach to baselines that follow the same architecture (Figure~\ref{fig:framework}b) but use different visual representations. As an upper bound, we consider oracle hand-engineered states for solving a task (e.g., 3D poses and direction-to-goal vectors). We also compare our method to visual encoders trained by supervised learning on ImageNet~\cite{Deng2009}, the choice of encoder in most vision tasks. We summarize our main results as follows:

- 1) We show that a \emph{single} visual encoder pre-trained on real-world images can solve various motor control tasks \emph{without} fine-tuning per-task, state estimation, or demonstrations.

- 2) Our \emph{self-supervised} approach consistently outperforms \emph{supervised} representations (up to 80\% absolute success rate), and even matches the oracle performance in some cases. 

- 3) We find that pre-training on \emph{images in the wild}, e.g., from YouTube~\cite{Shan2020} or Egocentric~\cite{Damen2018,Damen2021} videos, works better for manipulation tasks than ImageNet~\cite{Deng2009} images.

- 4) We show that our visual representations \emph{generalize} in various ways. For example, our visual encoder disentangles shape and color and is able to handle a range of different object geometries and configurations.

We encourage researchers to evaluate visual representations not only on downstream vision tasks but also on motor control tasks. We believe that our work is a promising step in this direction and release the benchmark suite, pre-trained models, and the training code on the \textbf{\href{https://tetexiao.com/projects/mvp}{project page}}.

\newpage

\section{Masked Visual Pre-training for MC}

\subsection{Masked Visual Pre-training}\label{sec:ssvp}

Thanks to the Internet and portable camera devices (e.g., phones, glasses, etc.), we now have access to large amounts of video data to learn from in various ways. We leverage such data for learning \emph{visual representations}. Specifically, we use images from the egocentric Epic Kitchens dataset~\cite{Damen2018,Damen2021}, the YouTube 100 Days of Hands dataset~\cite{Shan2020}, and the crowd-sourced Something-Something dataset~\cite{Goyal2017}. Combined, these sources yield a collection of $\app$700K images that we refer to as the Human-Object Interaction dataset (HOI). Note that we do \emph{not} exploit any human labels or temporal information even if it is possible. We also apply our approach using the ImageNet dataset~\cite{Deng2009} for controlled comparisons with supervised baselines.

With the data in hand, we must now formulate an appropriate self-supervised task. We adopt masked modeling as our self-supervision objective---specifically, we use masked autoencoder (MAE)~\cite{He2021}. MAEs mask-out random patches of the input image and reconstruct the missing pixels with a Vision Transformer (ViT)~\cite{Dosovitskiy2020}.  During training, only unmasked patches are fed into the MAE; this strategy makes training more efficient. It is critical to train MAE with a high masking ratio (e.g., masking 75\% of all patches) and use a heavy encoder with a light decoder. The most appealing property of MAE is its simplicity and minimal reliance on dataset-specific augmentation engineering; for example, it works well even with minimal data augmentations (center crop and color).

In Figure~\ref{fig:mae_vis} we show example reconstructions for HOI images. We observe that the model learns about color, shape, and objects. Notice that the images are representative of everyday interactions making them well suited for our needs.

\subsection{Learning Motor Control from Pixels}

Given the pre-trained visual encoder, we now turn to learning motor control from pixels. We freeze the visual representations and use them for all downstream motor control tasks; we do not perform any task-specific fine-tuning of the image encoder. This design has two main benefits. First, it prevents the encoder from overfitting to the setting at hand and thus preserves general visual representations for learning new tasks. Second, it leads to considerable memory and run time savings since there is no need to back-propagate through the encoder. Freezing visual encoder makes using large vision models in the RL loop fast and feasible.

In Figure~\ref{fig:framework}b, we show our architecture for learning motor control from pixels. We first extract a fixed-sized vector of image features using our pre-trained visual encoder. Notice that all of the image patches are passed through the encoder, unlike in the masked pre-training stage. We additionally compile proprioceptive robot information in the form of joint positions and velocities into a second vector. This proprioceptive information is readily available on real robot hardware. We concatenate these two vectors to obtain the input embedding for the neural network controller.

We then train task-specific motor control policies on top of this embedding with model-free reinforcement learning. Specifically, we use the proximal policy optimization (PPO) algorithm~\cite{Schulman2017}. PPO is a state-of-the-art policy gradient method that has shown excellent performance on complex motor control tasks and successful transfer to real hardware~\cite{Openai2018,Openai2019}. Our policy is a small multi-layer perceptron (MLP) network. In addition, we train a critic that has the same architecture as the policy using the same representations. The policy and the critic do not share weights.

\newpage

\begin{table}[t]\centering
\resizebox{\columnwidth}{!}{\begin{small}\begin{sc}\begin{tabular}{lcccc}
\toprule
           & RLBench       & Robosuite & MetaWorld  & Ours     \\
\midrule
Simulator        & Coppelia      & MuJoCo    & MuJoCo     & IsaacGym \\
Fast       &               &           &            & \checkmark  \\
\#Arms      & 1             & 8         &  1         &    2     \\
\#Hands      &               &           &            &    \checkmark     \\
\#Tasks      & 100           &   9       & 50         &   8     \\
Rewards     &               & \checkmark   & \checkmark    & \checkmark  \\
\bottomrule
\end{tabular}\end{sc}\end{small}}\vspace{-2mm}
\caption{\textbf{Existing benchmarks}. Compared to existing benchmarks, ours features a unique combination of hand-designed tasks, dense rewards, and complex robots (e.g., multi-finger hands). Crucially, it leverages a fast simulator and provides distributed training for scaling learning-based motor control from pixel observations.}
\label{tab:benchmarks}\vspace{-2mm}
\end{table}

\newpage

\section{\pmc Benchmark}

We construct a new benchmark suite of tasks for studying motor control from pixels, described in this section.

\subsection{Motivation}

While there exist a number of excellent benchmarks for motor control, e.g., DMC~\cite{Tassa2018}, RLBench~\cite{James2020}, Robosuite~\cite{Zhu2020}, MetaWorld~\cite{Yu2020}, they all fall short on one or more of our requirements. In particular, there is no benchmark suite for learning motor control algorithms that has high-resolution images, realistic robots, fast physics simulation, efficient training, and appropriate reward functions and metrics. To this end, we introduce a new benchmark suite for \emph{Pixel Motor Control}, which we call \pmc. We compare the key aspects of \pmc to several existing benchmarks in Table~\ref{tab:benchmarks}.

\subsection{Simulator}

We leverage the recent NVIDIA IsaacGym simulator~\cite{Makoviychuk2021} to build our benchmark. The core design idea of IsaacGym is to perform simulation on a GPU in a shared context. IsaacGym allows for very fast training times. For example, we are able to train our oracle state-based models in $\app$12 minutes and our image models in $\app$5 hours ($\app$8 million environment steps) on a single NVIDIA 2080 Ti GPU. This training speed is considerably faster than it would be in other simulators commonly used for motor control such as MuJoCo~\cite{Todorov2012}. \citet{Rudin2021} have shown that sim-to-real transfer is feasible based on policies trained in IsaacGym. 

\subsection{Robots}

\pmc includes two robot arms, a parallel jaw gripper, and a multi-finger hand combined as follows: \textbf{(1)} \emph{Franka:} A Franka Emika robot commonly used for research. It has a 7-DoF arm with a 2-DoF gripper. \textbf{(2)} \emph{Kuka with Allegro:} A Kuka LBR iiwa arm with 7 DoFs and a 4-finger Allegro hand with 16 DoFs (4 DoFs per finger), for 23 DoFs in total. For brevity, we refer to it as ``Kuka.''

\begin{figure}[t]\centering\vspace{-0mm}
\includegraphics[width=1.0\linewidth]{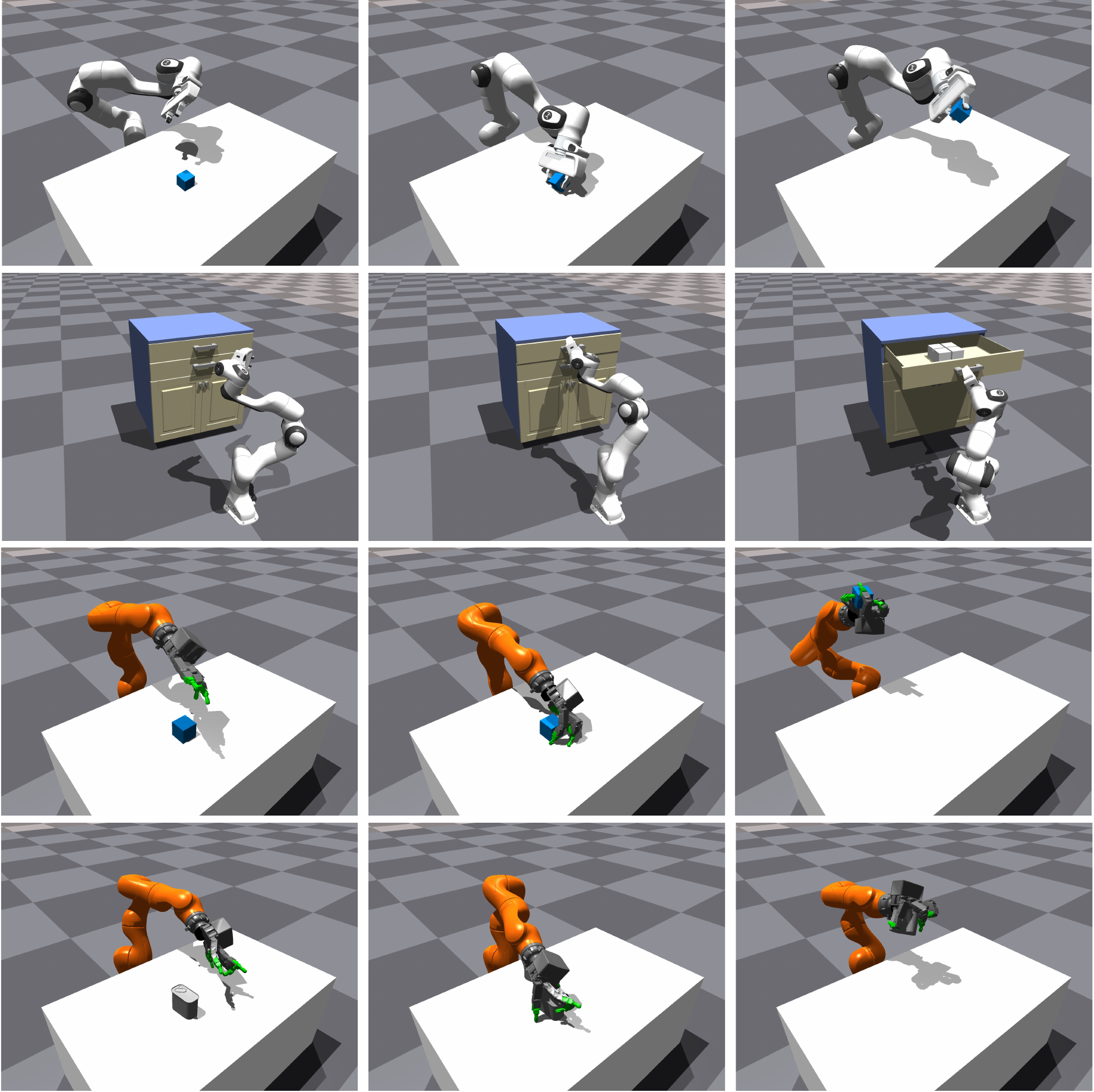}\vspace{-2mm}
\caption{\textbf{Example tasks.} We show example trajectories for the Franka and Kuka tasks. See the \textbf{\href{https://tetexiao.com/projects/mvp}{project page}} for video examples.}
\label{fig:benchmark}\vspace{-2mm}
\end{figure}

\subsection{Observations and Actions}

Our benchmark supports rendering high-resolution pixel observations for each robot. For both the Franka and Kuka setups and each of the tasks, we use wrist-mounted cameras by default. The benchmark provides proprioceptive information for the robots, as well as hand-engineered states typically including 3D poses or relevant objects, goals, and their relations. All of our default settings use position control in joint angle space with a control frequency of 60Hz.

\begin{figure*}[t]\centering
\includegraphics[width=1.0\linewidth]{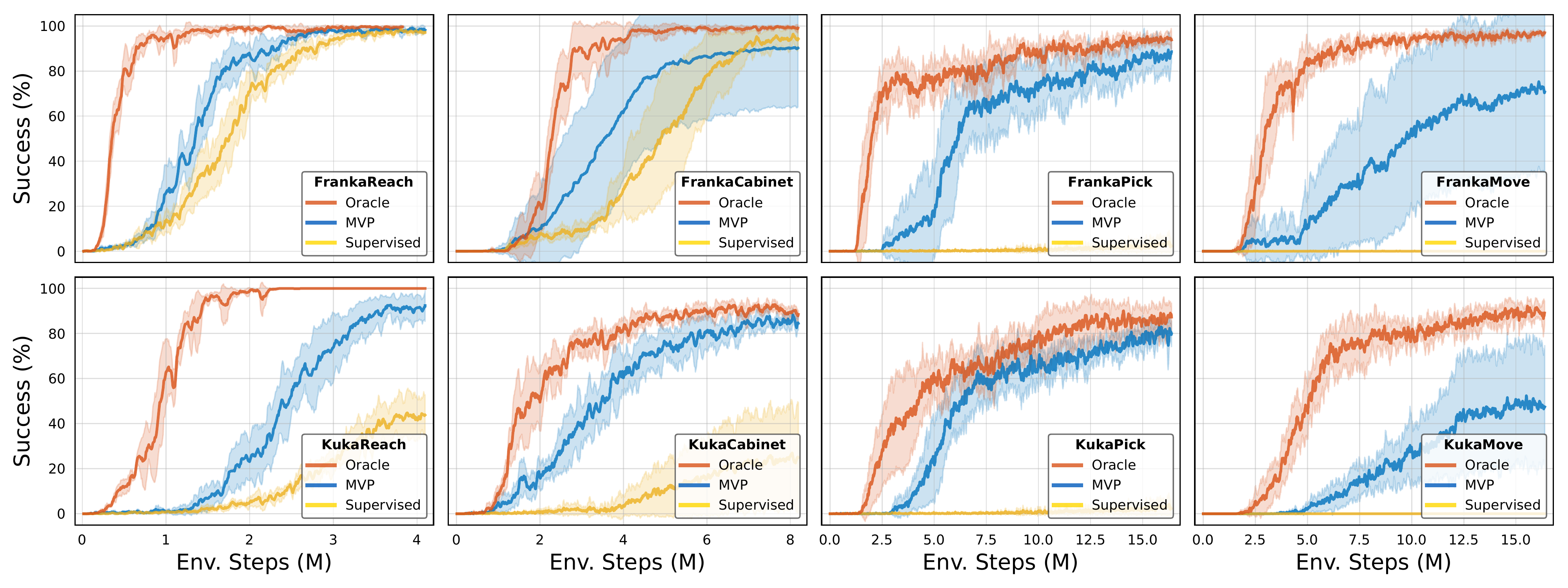}\vspace{-2mm}
\caption{\textbf{Sample complexity.} We plot the success rate as a function of environment steps on the 8 \pmc tasks. Each task uses either the Franka arm with a parallel gripper or the Kuka arm with a multi-finger hand. The \method approach significantly outperforms the supervised baseline on 7 tasks and closely matches the oracle state model (considered the upper bound of RL) on 5 tasks at convergence. The result shows that self-supervised pre-training markedly improves representation quality for motor control tasks.}
\label{fig:sample_complexity}\vspace{-0mm}
\end{figure*}

\subsection{Tasks, Rewards, and Metrics}

\pmc tasks include several movement types from basic reaching to interacting with objects. The objects in the environment vary in positions, scale, color, and shape. Figure~\ref{fig:teaser} and~\ref{fig:benchmark} show a few example scenes. We hand-design task-specific dense reward functions for training RL policies. We define reward-independent success metrics that typically quantify the distance from the agent or an object to a specified goal location over sufficient time steps.

\subsection{Distributed Training}

The scarcity of GPU memory is a bottleneck for learning motor control from pixels. For our typical setup with 224\x224 images, we can fit at most 256 environments on a single 2080 Ti GPU. We implement PPO with distributed training to support large batch sizes. Similar to data parallel training, we create a model replica per-GPU, collect rollouts on each GPU, and synchronize gradients across GPUs.

\section{Experimental Setup}


\paragraph{Data for pre-training.} We consider two kinds of pre-training data: ImageNet~\cite{Deng2009} and a joint Human-Object Interaction (HOI) dataset. We construct the HOI data by combining Epic-Kitchens~\cite{Damen2018,Damen2021}, Something-Something~\cite{Goyal2017}, and 100 Days of Hands (100-DOH)~\cite{Shan2020}. To build HOI, we sample frames from Epic-Kitchens and Something-Something at 1fps and 0.3fps, respectively. This yields 700k images including 100k from 100-DOH.

\paragraph{Encoder.}
The image encoder follows standard ViT architecture~\cite{Dosovitskiy2020}. ViT partitions an image into patches and linearly projects each patch into features, followed by standard Transformer blocks~\cite{Vaswani2017, Wang2019}. We use the ViT-Small model with a 16\x16 patch size, 384 hidden size, 6 attention heads, an MLP multiplier of 4, and 12 blocks. The model runs at 4.6 gigaflops for input images of 224\x224, approximately 1.2\x~as many as the ResNet-50~\cite{He2016} model. 

We pre-train supervised and self-supervised variants of the ViT model. For the supervised model, we use the recipe in~\cite{Xiao2021} and train on the ImageNet dataset for 400 epochs. We use the MAE framework for the self-supervised counterpart~\cite{He2021}. We use an auxiliary dummy classification token in the MAE for downstream finetuning and transfer~\cite{He2021}. We use a crop ratio of [0.2, 1.0] for ImageNet and [0.1, 0.75] for HOI, due to the larger width-over-height aspect ratio of HOI images. We train the MAE models for 1600 epochs on 16 GPUs for both HOI and ImageNet datasets.

\paragraph{Controller.}
The controller is a simple MLP with each hidden layer followed by a SeLU~\cite{Klambauer2017} activation function. We use a four-layer MLP with hidden layers of size [256, 128, 64] for all tasks, following~\cite{Makoviychuk2021}. The (dummy) classification token of the ViT encoder yields the image features and a linear layer projects the features to 128 dimensions. The controller takes in the linearly-projected (128-d) proprioceptive state of the robot along with the projected image features. The controller outputs delta joint angles.

\paragraph{Training with RL.}
We freeze the visual encoder throughout the entire training horizon. We train for 500 iterations for reach, 1000 iterations for cabinet, and 2000 iterations for pick and relocate, respectively. In each iteration, we collect samples from 256 environments which have 32 steps each. We train for 10 epochs on these collected samples per iteration. We compose 4 minibatches per epoch, i.e., 4 gradient updates, leading to a minibatch size of 2048 per gradient update. We choose this configuration because it maximizes the memory on a single NVIDIA 2080 Ti GPU. In all experiments we train with a cosine learning rate decay schedule~\cite{Goyal2017b}. To reduce randomness in the RL experiments~\cite{agarwal2021}, for each task and model we search for the best learning rate in \{0.0005, 0.001, 0.0015\} with 5 seeds per learning rate (15 runs for each task and model). We always report the performance yielded by the best learning rate aggregated over seeds unless otherwise specified. Other hyperparams use defaults: Adam optimizer with $\beta_1=0.9$ and $\beta_2=0.999$, gradient norm of 1, initial noise standard deviation of 1.0.

\section{Experimental Results}

\subsection{Sample Complexity}
Figure~\ref{fig:sample_complexity} shows success rates over training on 8 challenging tasks from \pmc. We consider the oracle state model (i.e., position, orientation, and velocity of the object, goal and robot in world-coordinate system, which is difficult to estimate in real-world settings) as the upper bound of RL. \method significantly outperforms the supervised baseline on 7 out of 8 tasks and matches the baseline on the 8th task. At convergence, \method closely matches the oracle state model on 5 tasks. The supervised baseline is flat at zero success rate on the pick and move tasks with both robots; \method rivals the oracle on the pick task and achieves high success rate on the relocate task. These results show that self-supervised pre-training markedly improves representation quality for motor control tasks.

\begin{figure}[t]\centering
\includegraphics[width=1.0\linewidth]{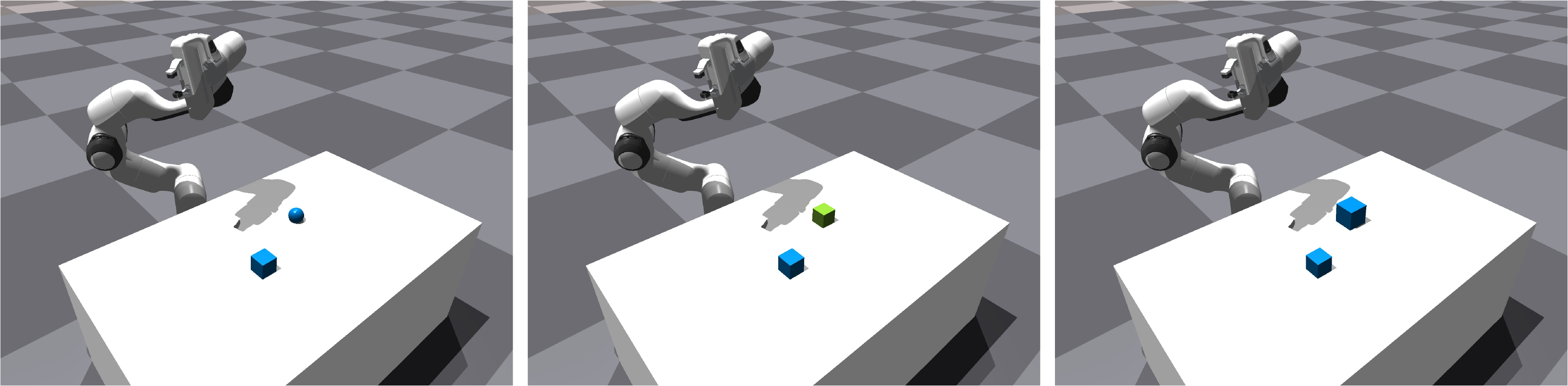}\\[2mm]
\includegraphics[width=1.0\linewidth]{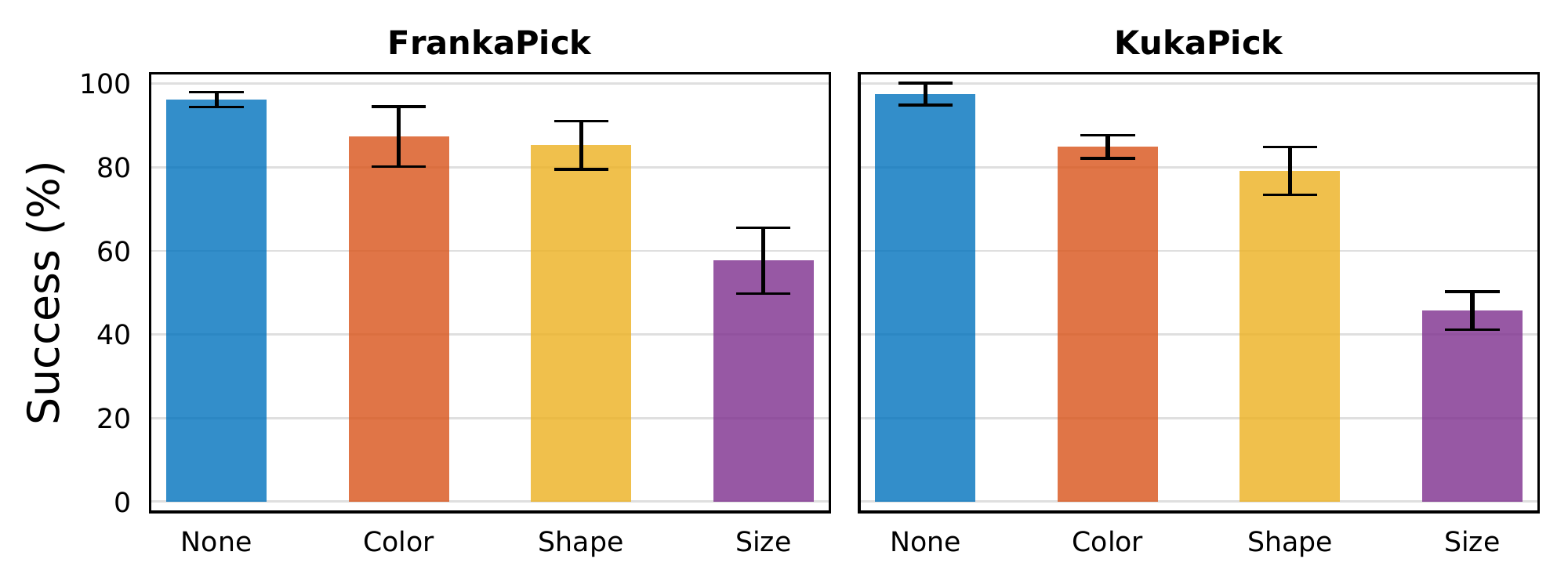}\vspace{-4mm}
\caption{\textbf{Robustness to distractors.} The robots are trained to pick up a blue box of 4.5cm side length. At \emph{test time}, we add a distractor object differing from the training object in terms of color (blue vs. green), shape (cube vs. sphere), or size (4.5cm vs. 6cm), shown at the top. \method maintains high success rates for color and shape. The model is less sensitive to size likely due to the scale ambiguity from single first-person camera setup.}
\label{fig:robustness}\vspace{-0mm}
\end{figure}

\subsection{Generalization}

We design various experiments on the pick task with both the Franka and Kuka arms to demonstrate the degree to which \method is able to generalize.

\begin{figure}[t]\centering
\includegraphics[width=1.0\linewidth]{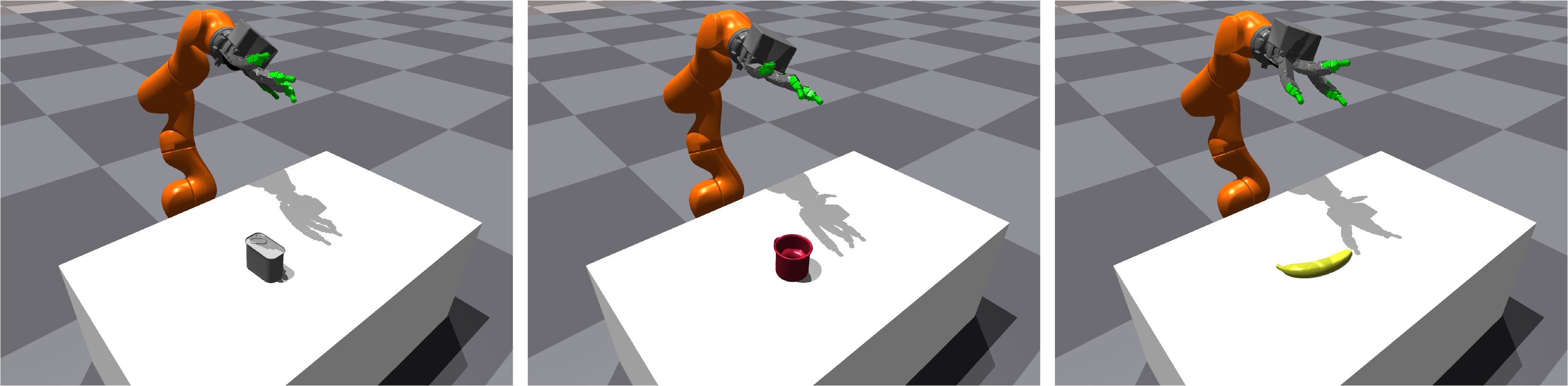}\\[4mm]
\includegraphics[width=1.0\linewidth]{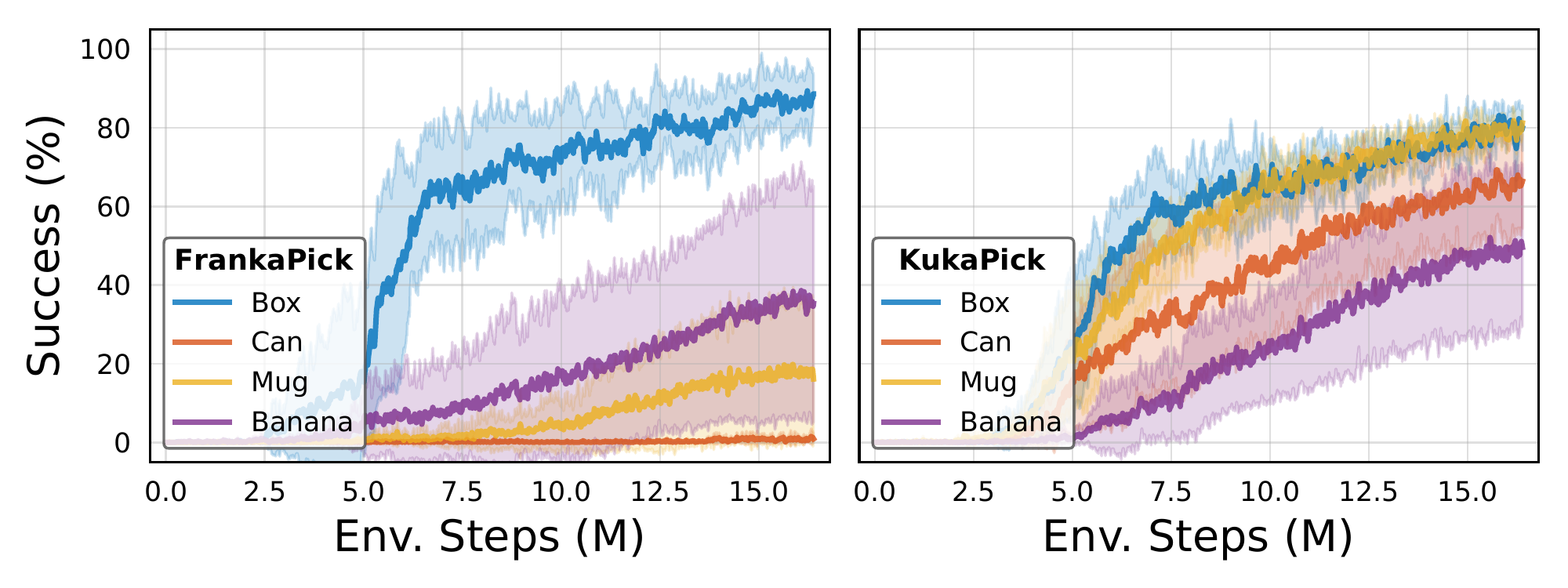}\vspace{-4mm}
\caption{\textbf{Generalization to objects of various geometries.} We import three additional objects (i.e., can, mug, and banana) from the YCB dataset and re-train the controller to pick up the individual object category. The Kuka robot with the Allegro hand can pick up all objects with $\app$50\% success rate.}
\label{fig:many_pick_objects}\vspace{-0mm}
\end{figure}

\paragraph{Robustness to distractors.}
In the pick task the robots are trained to pick up a blue box of 4.5cm side length. In this experiment we add a distractor object that differs from the training object in terms of color, shape, or scale, \emph{at testing time}. The models are not retrained for new testing configurations. A robust model from pixels should pick up the object used for training. Specifically, we have 1) a \emph{green} box of 4.5cm side length (color distractor); 2) a blue \emph{sphere} of 4.5cm diameter (shape distractor); and 3) a blue box of \emph{6cm side length} (scale distractor). Figure~\ref{fig:robustness} shows the results of our \method model. Color and shape distractors only marginally decrease the success rate, implying that \method is able to recognize the color and shape of objects. The model, however, is less sensitive to scale variation as the 50\% success rate suggests that the distractor or the original box is picked up by chance. We believe it is due to scale ambiguity from single first-person camera setup. Note that the oracle state model would not be able to function in these testing cases due to changes in state dimensions. 

\paragraph{Generality of the framework.} We import various objects from the YCB dataset~\cite{Calli2015}---box, can, mug, and banana---for the pick task and re-train the model for each individual object. Figure~\ref{fig:many_pick_objects} visualizes the experiment setup and shows the results. The Kuka robot with the Allegro hand can pick up all of the objects with at least a 50\% success rate. This shows \method as a framework can generalize to objects of different geometries and the multi-finger hand's strength in object manipulation. 

\begin{figure*}[t]\centering
\includegraphics[width=1.0\linewidth]{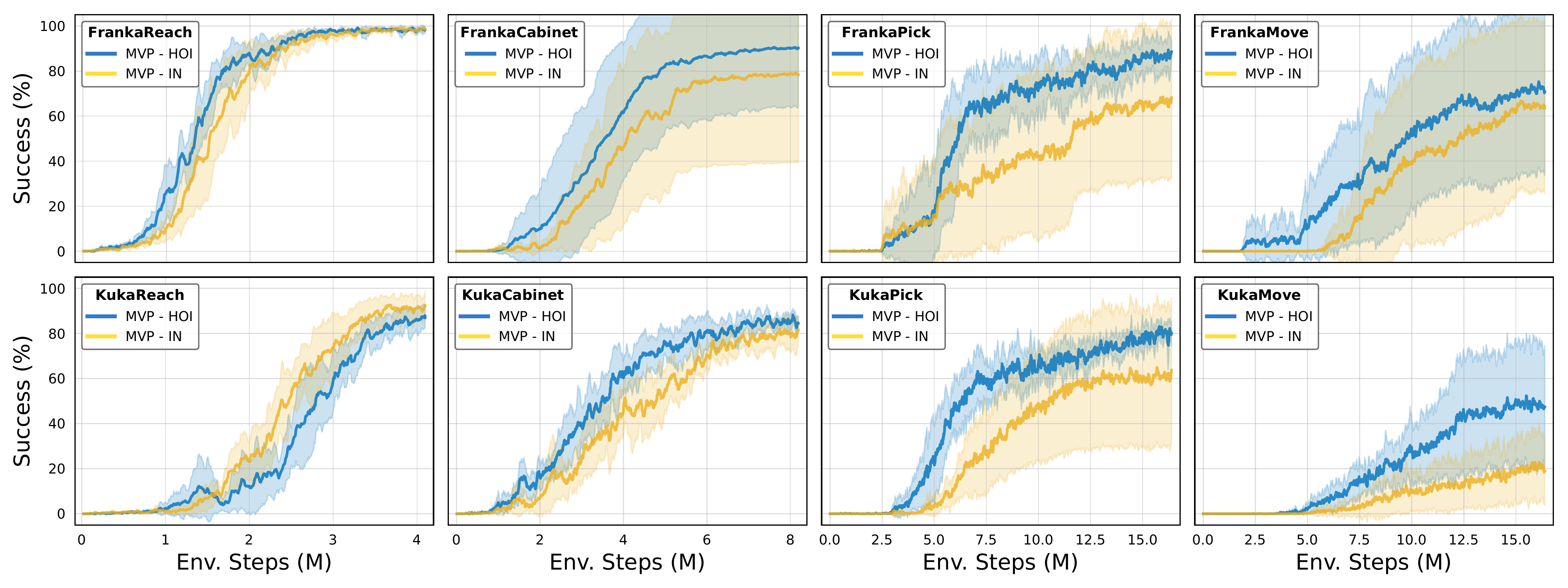}\vspace{-4mm}
\caption{\textbf{Pre-training data: HOI vs. ImageNet.} We compare the performance of our approach when using HOI and ImageNet images as the source of pre-training data. Overall, we find that the representations learned from HOI data perform better on motor control tasks.}
\label{fig:data_comparison}\vspace{-0mm}
\end{figure*}

\subsection{Ablations}

\paragraph{Pre-training data.}
We train \method on HOI and ImageNet data, respectively. Figure~\ref{fig:data_comparison} shows the results. \method trained on HOI data outperforms the counterpart trained on ImageNet data on 7 out of 8 tasks. Whereas ImageNet is dominated by images of animals and objects, HOI contains many images demonstrating object manipulation from a first-person camera view. We hypothesize that this difference is why HOI is empirically the superior choice for motor control tasks.

\begin{figure}[t]\centering
\includegraphics[width=1.0\linewidth]{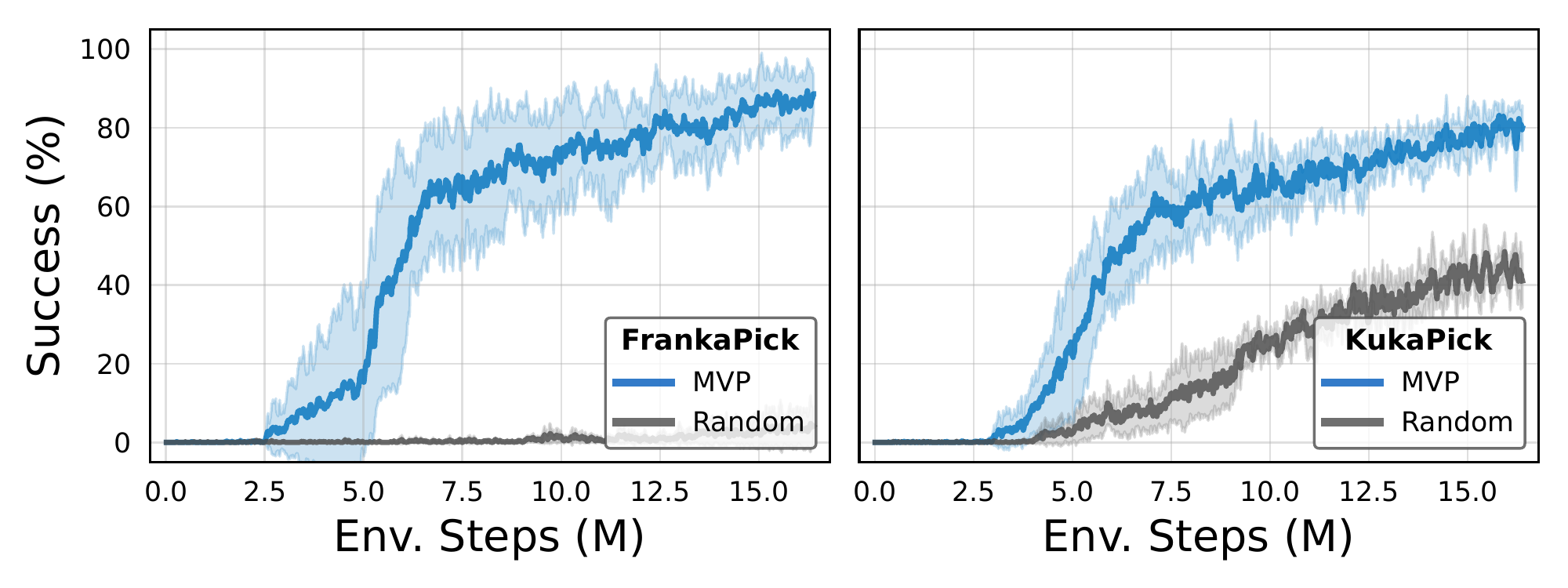}\vspace{-4mm}
\caption{\textbf{Random features.} We compare our learnt representations to a random features baseline. We use the same visual encoder, initialize it randomly, and freeze. The random model fails on 6 out of 8 \pmc tasks (0 success rate). Here we show one task that fails (FrankaPick) and one that succeeds (KukaPick).}
\label{fig:random_feature}\vspace{-2mm}
\end{figure}

\paragraph{Random features.}
We compare our \method with a randomly initialized and frozen image encoder. The random model yields flat-zero or close-to-zero success rates on 6 out of 8 tasks. This indicates that random features are insufficient to solve complex motor control tasks. We show the results of one successful and one failed task in Figure~\ref{fig:random_feature}.

\begin{figure}[t]\centering
\includegraphics[width=1.0\linewidth]{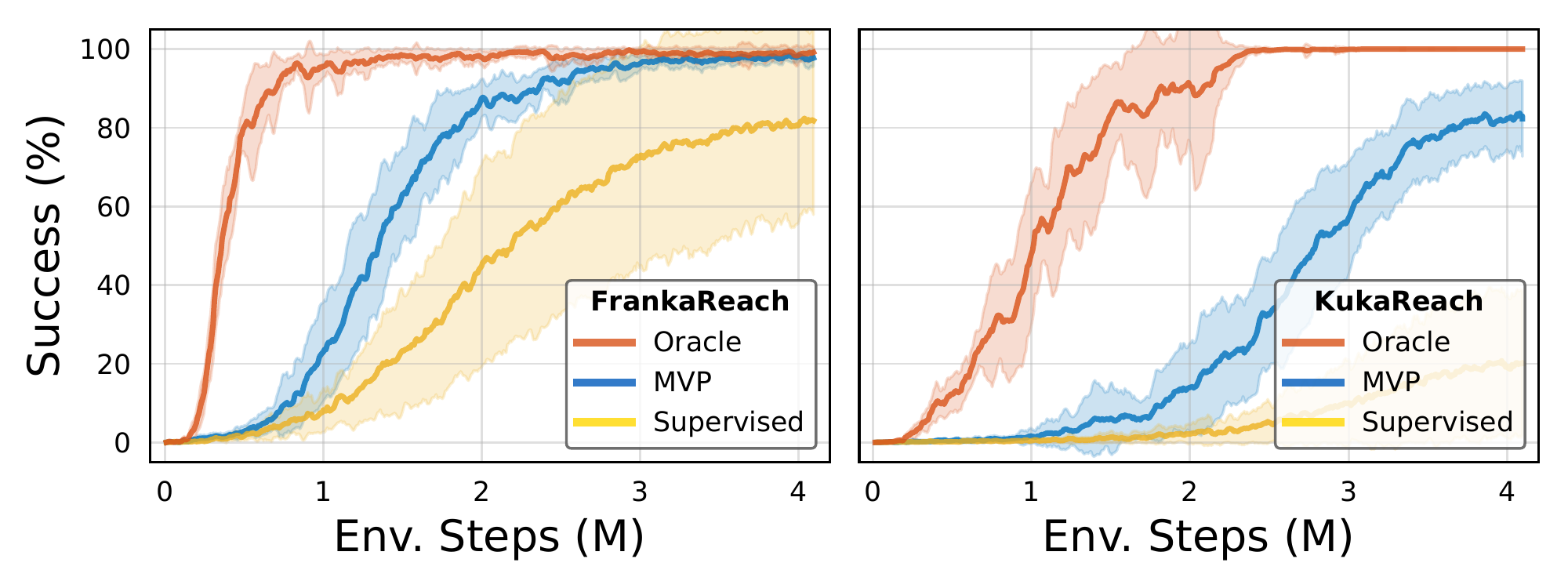}\vspace{-4mm}
\caption{\textbf{Learning rate and seed stability}. For each model, we train 15 instances of the model with 3 learning rates and 5 seeds. We show the results on tasks where the supervised baseline achieves nontrivial performance. \method significantly improves stability over the supervised model.}
\label{fig:stability}\vspace{-2mm}
\end{figure}

\paragraph{Stability.}
We characterize how sensitive different models are to changes in learning rate and random seed. Representations of high quality should be less sensitive and yield consistently-high performance. Figure~\ref{fig:stability} shows the reach task results over 3 learning rates (0.0005, 0.001, 0.0015) and 5 seeds for a total of 15 runs for each model. Although the supervised baseline with the best choice of learning rate performs close to \method in Franka reach and has over 40\% success rate in Kuka reach (see Figure~\ref{fig:sample_complexity}), it exhibits much worse stability across a range of learning rates. Our \method, in contrast, shows good stability in the learning rate hyperparameter, a sign of superior optimizability in training.

\begin{figure}[t]\centering
\includegraphics[width=1.0\linewidth]{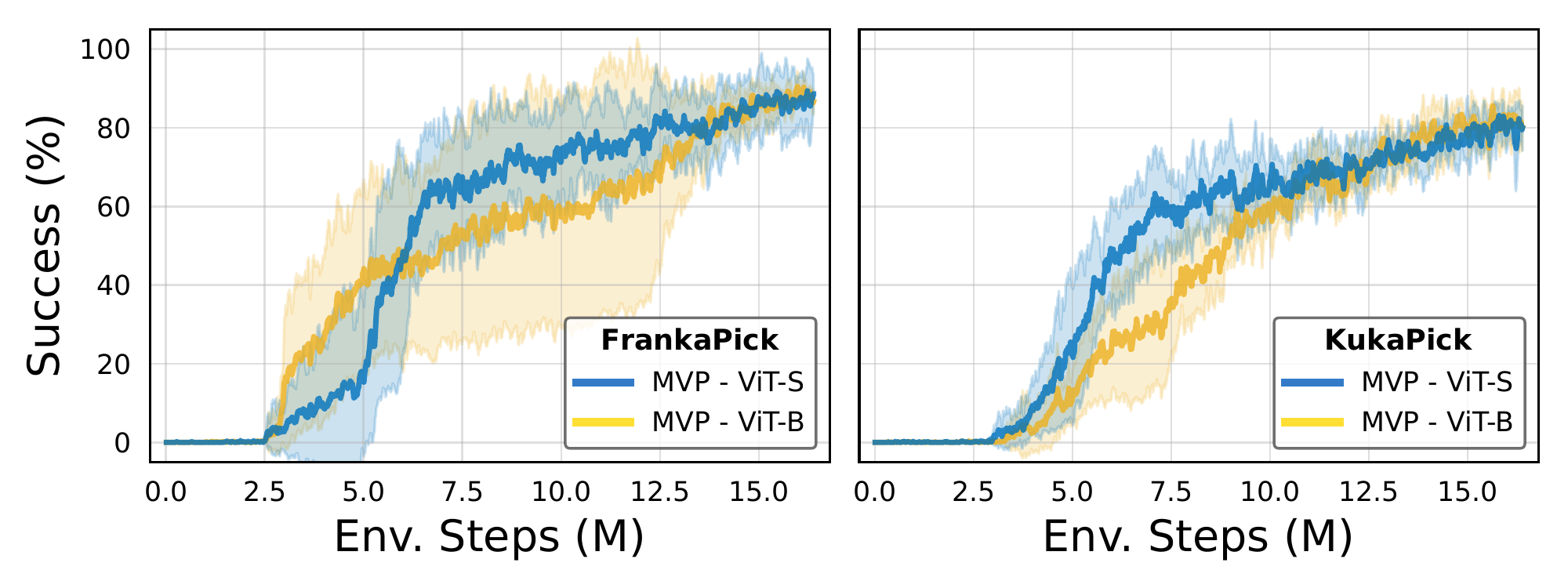}\vspace{-4mm}
\caption{\textbf{Larger encoders.} We pre-train a ViT-Base model (18Gflops) and use the representations for the pick task. We do not observe clear gains from preliminary model scaling and believe that scaling data and model size is an exciting area for future work.}
\label{fig:scaling}\vspace{-2mm}
\end{figure}

\begin{figure}[t]\centering
\includegraphics[width=1.0\linewidth]{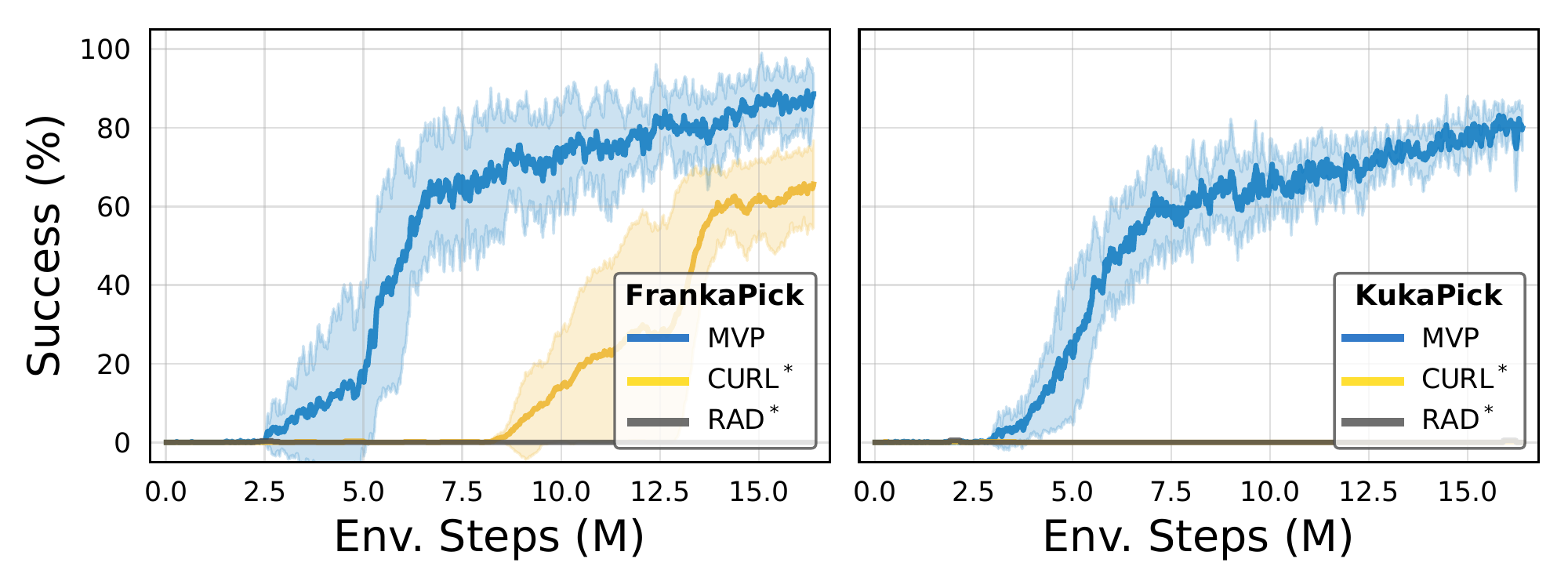}\vspace{-4mm}
\caption{\textbf{CURL and RAD comparisons.} We compare \method to environment training with our reimplementation of CURL and RAD (denoted with star; see text for details). We observe that, for a fixed number of steps, \method outperforms both while being less computationally expensive to train (1 vs. 8 GPU training).}
\label{fig:curl}\vspace{-2mm}
\end{figure}

\paragraph{Unfreezing encoders.} We experiment with training the encoder end-to-end with RL in downstream motor control tasks. Unfreezing the encoder significantly increases GPU compute and memory consumption. To maintain the number of environments, we increase the number of GPUs used from 1 to 8 via our distributed training pipeline. We choose the Franka pick task as the testbed and run 3 runs with different seeds per model. We test with two initialization: (1) initializing the visual encoder randomly, or (2) initializing with the pre-trained MAE weights. We observed unstable training (the loss goes to NaN), and we decreased the learning rate until training successfully completed. Still, both models yielded flat zero success rate on the task at all seeds. This result is somewhat counter-intuitive as one would expect end-to-end training should yield better results than training on frozen representations. We conjecture that it may be due to (a) the RL signal being unstable and hard to tune; (b) noisy gradients from the RL objective interfering with pre-trained visual representation; or (c) a high capacity vision model like ViT requiring significantly more samples and environments to train end-to-end with RL. Freezing the visual encoder as in \method preserves the quality of visual representations while yielding faster RL training.

\paragraph{Larger encoders.} We pre-train a ViT-Base encoder (18 gigaflops) and conduct a preliminary transfer study on the Franka/Kuka pick tasks. Figure~\ref{fig:scaling} shows the results. We observe that the larger encoder does not improve performance. A larger encoder potentially requires more data and/or a different training recipe. Overall, scaling data and models in the context of self-supervised representations for motor control remains an exciting area for future work.

\subsection{Additional Comparisons}

\paragraph{CURL and RAD comparisons.} 
We compare \method to two state-of-the-art methods that train the vision encoder with environment data: CURL~\cite{Srinivas2020} and RAD~\cite{Laskin2020}. Due to differences in their original settings and ours, e.g., small ConvNet vs. large ViT, we compare to our reimplementations (denoted with star). In particular, we adopt PPO, ViT-Small visual encoder, and MoCo-v3~\cite{Chen2021iccv} data augmentation recipe. We observe that, for a fixed number of steps, \method outperforms both baselines while being less computationally expensive (1 vs.\ 8 GPUs). We note that environment training with enough data might yield better performance than our image pre-training. Indeed, we do not see our approach as the ultimate answer for any one setting but rather as a solid baseline, or a starting point, for many varying settings.

\newpage

\begin{figure}[t]\centering
\includegraphics[width=1.0\linewidth]{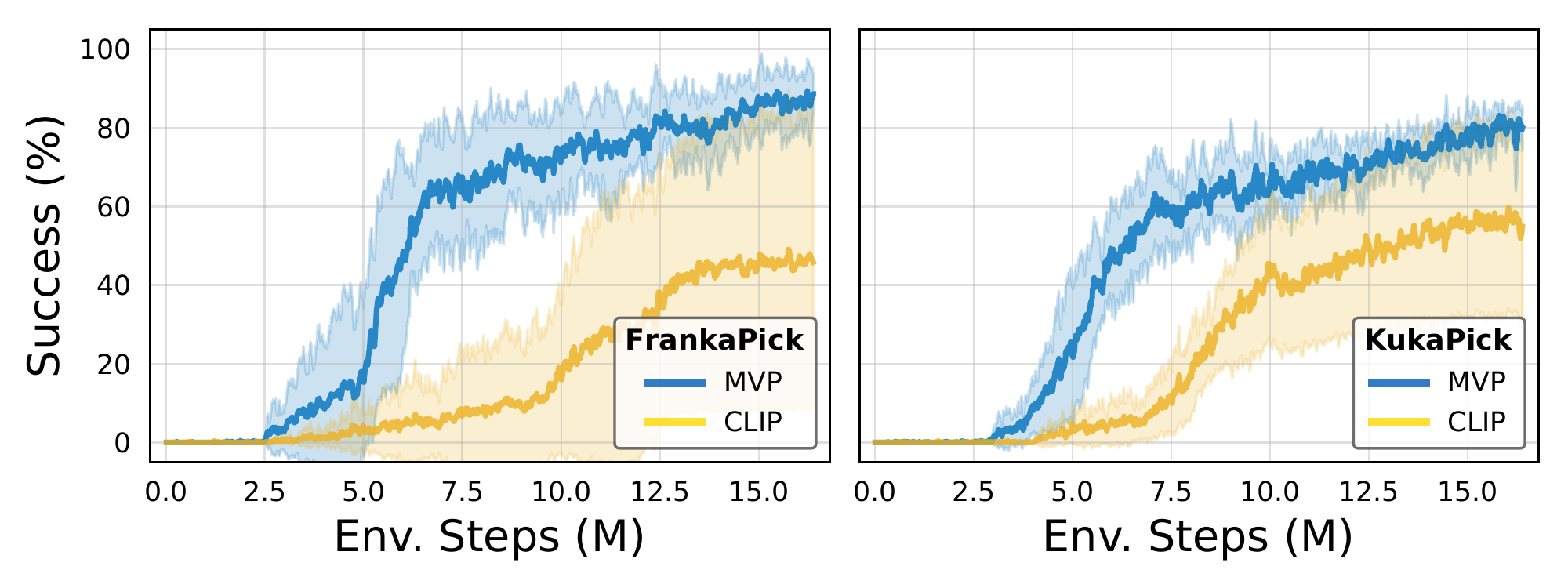}\vspace{-4mm}
\caption{\textbf{CLIP comparisons.} We compare our visual representations to the CLIP visual encoder trained with large scale language supervision. The results show a promising signal that self-supervised representations can outperform CLIP encoders.}
\label{fig:clip}\vspace{-2mm}
\end{figure}

\paragraph{CLIP comparisons.} Next, we experiment with substituting a pre-trained CLIP visual encoder~\cite{Radford2021} in place of our MAE encoders. The CLIP encoder is trained using 400M labeled text-image pairs and has shown excellent performance across a wide range of visual tasks. We opt to use the ViT-Base CLIP encoder as it is closest in size to our ViT-Small encoders. We note that this comparison is imperfect due to difference in data distributions and encoders but believe it is still instructive to see. In Figure~\ref{fig:clip} we show the results. We observe a promising signal that self-supervised representations can outperform strong CLIP encoders.
We believe it would be interesting to perform a controlled study on in-the-wild images with text annotations, like from the recently released Ego4D dataset~\cite{Grauman2021}.

\begin{figure}[t]\centering
\includegraphics[width=1.0\linewidth]{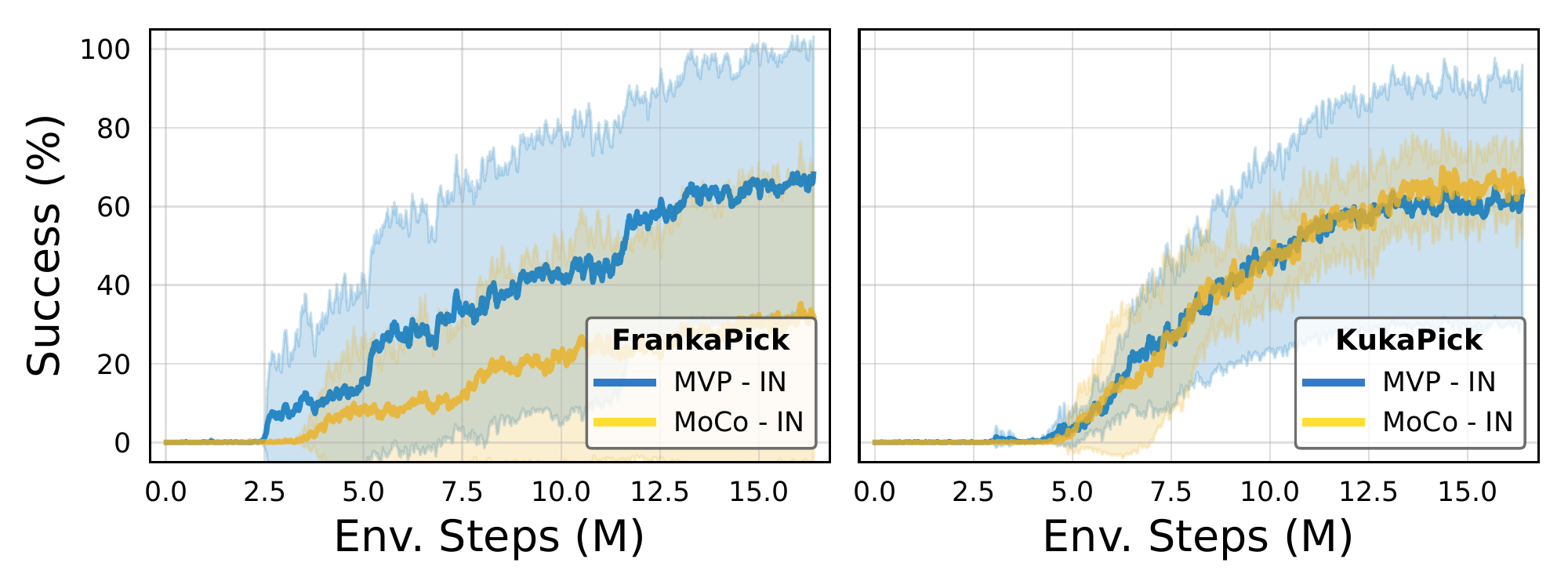}\vspace{-4mm}
\caption{\textbf{MoCo-v3 comparisons.} We compare MAE, used in \method, to the MoCo-v3 alternative for self-supervised pre-training \emph{on ImageNet}. We see that MoCo-v3 can achieve non-trivial performance, showing the generality of our approach. However, in contrast to MAE, it may be harder to adapt to in-the-wild images.}
\label{fig:moco}\vspace{-2mm}
\end{figure}

\paragraph{MoCo-v3 comparisons.} Finally, we compare visual encoders trained with the Momentum Contrastive (MoCo) self-supervised learning framework instead of MAE used in \method. We opt to use the latest MoCo-v3~\cite{Chen2021iccv} designed for ViT models. We show results in Figure~\ref{fig:moco}. We observe that MoCo-v3 can achieve good performance on one of the tasks and non-trivial on the other. This suggests that our approach may be more general and applicable to other self-supervised pre-training techniques as well. However, in contrast to MAE, it may be harder to adapt techniques like MoCo-v3 to in-the-wild images (see Figure~\ref{fig:data_comparison}).

\newpage

\section{Related Work}

\paragraph{Dexterous manipulation.} Recently, OpenAI~\cite{Openai2018, Openai2019} has shown impressive results in dexterous in-hand manipulation with large scale domain randomization. A number of works consider related manipulation problems with multi-finger hands but rely on explicit state estimation~\cite{Handa2020, Huang2021}, expert policies~\cite{Chen2021corl}, human demonstrations~\cite{Rajeswaran2018, Radosavovic2021, Qin2021}, human priors~\cite{Mandikal2021}, or models~\cite{Nagabandi2019}. In contrast, we do not use on any of the aforementioned components in our approach.

\paragraph{Representations in RL.} One way to learn representations for motor control is to rely on the task signal. This is commonly done in end-to-end RL~\cite{Mnih2015,Levine2016,Kalashnikov2018}. However, it results in high sample complexity, particularly in the case of high-dimensional observations like images. Furthermore, such representations may get overly adapted to the problem at hand and not generalize to new settings (e.g., new objects).

\paragraph{RL with self-supervision.} One way to overcome the high sample complexity of RL is to employ auxiliary objectives. In particular, in addition to learning the task, learn to predict some property of the environment, e.g., depth, that may lead to learning good representations as a side effect~\cite{Mirowski2017, Jaderberg2016, Shelhamer2017, Lample2017}. Rather than predicting hand-designed environment properties, researchers explored using more general self-supervised objectives~\cite{Oord2018, Yarats2019, Srinivas2020}. For example, \citet{Srinivas2020} show excellent performance in vision-based RL tasks. Representations can also be pre-trained on the data from the environment~\cite{Ha2018,Srinivas2020}. Overall, we share the goal of learning good visual representations with self-supervision. In contrast, we learn representations from large collections of natural images rather than environment-specific experience.

\paragraph{Self-supervision in robotics.} Self-supervised learning has also been used in various robotic settings. \cite{Pinto2016} and \citet{Agrawal2016} learn representations through interaction. \citet{Sermanet2018} learn representations from multiview video using contrastive learning and use them for imitation learning. \citet{Florence2018} learn dense image descriptors with self-supervision. \citet{Zhan2020} learn robotic manipulation with RAD~\cite{Laskin2020} and CURL~\cite{Srinivas2020}. \citet{Pari2021} show the effectiveness of visual representations with non-parametric nearest neighbor controllers.  All of these approaches learn representations in a specific robotic setting of interest (e.g., videos in the lab), rather than general visual representations from image collections like ours.

\paragraph{Supervised pre-training.} \citet{Sax2018} and \citet{Chen2020robust} show that representations learned from performing a set of mid-level vision tasks using label supervision benefits downstream navigation and manipulation tasks, respectively. \citet{Zhou2019} show the effectiveness on visual representations in driving settings. \citet{Yen2020} transfer image models trained on supervised vision tasks, e.g., edge detection and semantic segmentation, to affordance prediction models for object manipulation. \citet{Shah2021} use ImageNet representations for dexterous manipulation. In contrast to all of these, our approach is self-supervised and does not rely on labeled datasets.

\paragraph{Self-supervised pre-training in computer vision.} Self-supervised learning has been gaining momentum in computer vision. The approaches often rely on pretext tasks for pre-training~\cite{Doersch2015, Wang2015, Noroozi2016, Zhang2016, Pathak2016, Komodakis2018}. More recently, contrastive learning methods, e.g., \cite{Hadsell2006, Oord2018, Wu2018, Henaff2020, He2020, Chen2020,Jabri2020}, have been popular. These techniques try to learn to be invariant to a set of hand-crafted augmentations. \citet{Xiao2021a} have shown that the augmentations introduce inductive bias and may harm downstream transfer. Masked image autoencoding~\cite{Chen2020generative,Bao2021,He2021} pursues a different direction by learning to recover masked pixels. Specifically, we adopt the Masked Autoencoders (MAE)~\cite{He2021} have shown excellent performance on recognition tasks. We adopt the MAE as our visual pre-training strategy for learning motor control.

\section{Conclusion}

In this paper, we show that self-supervised visual pre-training is effective for motor control. We use a single vision encoder to learn various motor control tasks from pixels, without per-task fine-tuning, explicit state estimation, or expert demonstrations. We further show large sample complexity improvements compared to supervised baselines (up to 80\% absolute success rate) and sometimes even match the oracle state performance. Finally, we show that in-the-wild images, e.g., from YouTube or Egocentric videos, can lead to better visual representations than ImageNet images.

\section*{Acknowledgements}

We thank William Peebles, Matthew Tancik,  Anastasios Angelopoulos, Aravind Srinivas, and Agrim Gupta for helpful discussions. This work was supported in part by DOD including DARPA's MCS, XAI, LwLL, and/or SemaFor programs; ONR MURI program (N00014-14-1-0671), as well as BAIR's industrial alliance programs.

\clearpage

\bibliography{references}
\bibliographystyle{icml2022}

\end{document}